\documentclass[letterpaper]{article}
\usepackage{times}
\usepackage{url}
\usepackage{latexsym}

\usepackage{graphicx}
\usepackage{amsmath}
\usepackage{pst-all} 
\usepackage{com.braju.graphicalmodels} 
\usepackage{com.braju.pstricks} 
\usepackage{color}
\usepackage{hyperref}
\catcode`\@=11%

\title{An Adaptation of Topic Modeling to Sentences}
\author{
  Ruey-Cheng Chen\footnote{National Taiwan University.  email: {\tt rueycheng@ntu.edu.tw}},
  Reid Swanson\footnote{University of Southern California.  email: {\tt swanson@ict.usc.edu}},
  and Andrew S. Gordon\footnote{University of Southern California.  email: {\tt gordon@ict.usc.edu}},
}

\date{February 10, 2010}
\begin{document}
\maketitle

\begin{abstract}

Advances in topic modeling have yielded effective methods for characterizing
the latent semantics of textual data.  However, applying standard topic
modeling approaches to sentence-level tasks introduces a number of challenges.
In this paper, we adapt the approach of latent-Dirichlet allocation to include
an additional layer for incorporating information about the sentence boundaries
in documents.  We show that the addition of this minimal information of
document structure improves the perplexity results of a trained model.  

\end{abstract}

\section{Introduction} \label{s:introduction}

Topic models, such as probabilistic latent-semantic analysis
\cite{hofmann1999probabilistic} and latent-Dirichlet allocation
\cite{blei2003latent}, were first introduced in the NLP community as a means of
characterizing the latent semantics in textual data.  A large body of
research has made use of many successful variants of the
original models.  Generally, most of these previous efforts began with the
assumption that a document is, in principle, a bag of topics, and words 
observed in the document are indirectly inferred from the underlying
distribution of these background topics.  

The generative process for realizing a collection of words from a set of topics
relies on two distributions: the document-level topic distribution $\Pr(z|d)$
and the topic-level word distribution $\Pr(w|z)$, for topic $z$, document $d$
and word $w$.   This approach has garnered great success in various
document-based tasks.  However, there are problems in directly applying this
approach in sentence-level tasks, e.g., clustering individual sentences by
topic.

One way to apply LDA to sentence data is to treat individual sentences as whole
documents.  The problem with this approach is that it does not account for the
context in which the individual sentences appear.  Alternatively, the topic
distribution of individual sentences could be treated as the same as that of
the whole document.  Here the problem is that the differences between that
contributions of individual sentences to document topics is ignored.  Given the
large number of NLP applications where topic modeling could be applied to
sentence-level tasks, a new model is needed.

This work adapts the standard LDA model to better account for the contribution
of sentences to the topics of documents.  Our approach is to add an additional
layer in the standard LDA model that integrates information about the sentence
boundaries in documents.  The ensembles in this layer are indicators that point
to a set of multinomial distributions over topics; each sentence in the
document chooses a distribution to follow and generates the word topics
accordingly.  The idea is to offer a set of \emph{switches} in between
documents and topics, serving to balance the contribution of sentences and
documents to the topic distribution.  

In this paper, we present this new model as an adaptation of the standard
LDA model.  We demonstrate that our model performs better on test-set
perplexity than the standard LDA model.  These results suggest that this model
may have applicability in future sentence-level tasks.

\section{Sentence-Layered LDA} \label{s:sentence-layered-lda} 

\begin{figure}[ht!]
  \centering
  \psset{xunit=10mm,yunit=10mm}
  \begin{pspicture}(0,0)(8,9)
    \SpecialCoor  
    \psset{arrowscale=1.5}

    \rput(5.0,8.0){\GM@parameter{alpha}}\GM@label[angle=45]{alpha}{$\alpha$}
    \rput(2.0,8.0|alpha){\GM@node{theta}}\GM@label[angle=45]{theta}{$\theta$}
    \rput(theta|0,6.0){\GM@node{x}}\GM@label[angle=45]{x}{$x$}
    \rput(theta|0,4.0){\GM@node{z}}\GM@label[angle=45]{z}{$z$}
    \rput(theta|0,2.0){\GM@node[observed=true]{w}}\GM@label[angle=45]{w}{$w$}
    \rput(1.25,1.25){\GM@plate{1.5}{3.5}{$N$}}
    \rput(0.75,0.75){\GM@plate{2.5}{6.0}{$M$}}
    \rput(0.25,0.25){\GM@plate{3.5}{8.5}{$L$}}
    
    \rput(7.0,0|w){\GM@parameter{beta}}\GM@label[angle=45]{beta}{$\beta$}
    \rput(7.0,0|z){\GM@parameter{gamma}}\GM@label[angle=45]{gamma}{$\gamma$}
    \rput(5.0,0|w){\GM@node{phi}}\GM@label[angle=45]{phi}{$\phi$}
    \rput(5.0,0|z){\GM@node{tau}}\GM@label[angle=45]{tau}{$\tau$}
    \rput(4.25,3.25){\GM@plate{1.5}{1.5}{$S$}}
    \rput(4.25,1.25){\GM@plate{1.5}{1.5}{$T$}}

    \ncline[arrows=->]{alpha}{theta}
    \ncline[arrows=->]{theta}{x}
    \ncline[arrows=->]{x}{z}
    \ncline[arrows=->]{z}{w}
    \ncline[arrows=->]{beta}{phi}
    \ncline[arrows=->]{gamma}{tau}
    \ncline[arrows=->]{tau}{z}
    \ncline[arrows=->]{phi}{w}
  \end{pspicture}
  \caption{The sentence-layered LDA model in plate notation.  $S$ and $T$
  denotes the number of sentence and word topics; $N$, $M$, and $L$ denotes the
  number of words, sentences, and documents, respectively.}
  \label{plate-notation:sentence-layered}
\end{figure}
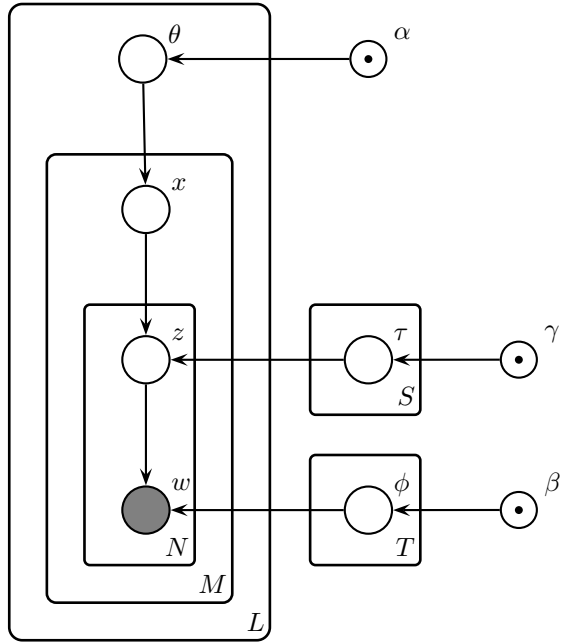

We propose a generative topic model, called {\it sentence-layered LDA}, that
incorporates sentence boundaries into the original LDA framework.  We introduce
the notion of {\it sentence topics} by adding a set of latent variables which
serve as additional sub-document constructs in between the document and the
words.  In this model, documents do not explicitly generate word topics, but
instead guide sentences toward certain topic distributions by producing a set
of sentence topics.  The sentence topics are indicator scalars, pointing to
specific discrete distributions over word topics.  The adaptation can be seen as
a number of LDA machines working as individuals at the sentence level being
governed by the document node.  The generative process is summarized by the
following variables.  
\begin{eqnarray*}
  \theta^{d} &\sim& \mathrm{Dirichlet}(\alpha) \\
  \tau^{(x_j)} &\sim& \mathrm{Dirichlet}(\gamma) \\
  \phi^{(z_i)} &\sim& \mathrm{Dirichlet}(\beta) \\
  x_j &\sim& \mathrm{Multinomial}(\theta^{(d)}) \\
  z_i &\sim& \mathrm{Discrete}(\tau^{(x_j)}) \\
  w_i &\sim& \mathrm{Discrete}(\phi^{(z_i)})
\end{eqnarray*}
Besides the usual constructs inherited from the LDA model, the newly-introduced
discrete distributions $\tau^{(x_j)}$ are governed by the Dirichlet prior
$\gamma$.   For simplicity, symmetric Dirichlet priors are assumed here.  The
complete plate notation of the model is shown in
Figure~\ref{plate-notation:sentence-layered}.

Since the joint probability $\Pr(\mathbf{z}, \mathbf{x})$ is generally
intractable, we break it down to two conditional probability distributions and
integrate out the priors.  The procedure largely follows the paradigm suggested
in \cite{blei2003latent} and \cite{griffiths2004finding}.  The inference is
done by a straightforward application of Gibbs sampler.

\paragraph{Conditional for $\mathbf{z_i}$.}  The conditional distribution $\Pr(z_i
= j|\mathbf{z}_{-i}, \mathbf{w}, \mathbf{x})$ can be shown to be as follows.
\begin{eqnarray*}
  && \Pr(z_i = j|\mathbf{z}_{-i}, \mathbf{w}, \mathbf{x}) \nonumber\\
  && \propto \Pr(w_i|z_i = j, \mathbf{z}_{-i}, \mathbf{w}_{-i}) \Pr(z_i = j|\mathbf{z}_{-i}, x, \mathbf{x}') \nonumber \\
  && = \frac{\beta + n_{-i,j}^{(w_i)}}{V \beta + n_{-i,j}^{(\cdot)}} \frac{\gamma + n_{-i,x}^{(j)}}{T \gamma + n_{-i,x}^{(\cdot)}} \label{z_i:2.3}
\end{eqnarray*}

Here, $x$ denotes the sentence topic that induces $z_i$, $\mathbf{x}$ denotes
the entire sentence set, and $\mathbf{x}' = \mathbf{x} - \{ x \}$.  The count
{for} all the occurrences for word $w$ that are of word topic $j$, excluding
the current assignment of at $i$-th word, is represented as $n_{-i,j}^{(w)}$;
the shorthand $n_{-i,j}^{(\cdot)}$ is a summation of the counts for all the
words.  In the second term, $n_{-i,x}^{(j)}$ denotes the number of words being
assigned to topic $j$ that are also governed by sentences assigned to the same
topic as that of $x$, with the current assignment at $i$-th word left out.  In
other words, we consider only sentences that are of same sentence-level topic
as that of $x$ and count the number of words in these sentences being assigned
to topic $j$.  

\paragraph{Conditional for $\mathbf{x}$.}  The conditional $\Pr(x =
l|\mathbf{x}',\mathbf{z})$ can be rewritten as: 

\begin{eqnarray*}
  && \Pr(x = l|\mathbf{x}',\mathbf{z}) \nonumber\\
  && \propto \Pr(\mathbf{z}^*|x = l, \mathbf{x}', \mathbf{z}') \Pr(x = l |\mathbf{x}',\mathbf{z}') \\
  && = \frac{\mathcal{B}(\{\gamma + n_{-x,l}^{(z)} + n_x^{(z)}; \forall z\})}{\mathcal{B}(\{\gamma + n_{-x,l}^{(z)}; \forall z\})}
  \frac{\alpha + n_{-x,d}^{(l)}}{S \alpha + n_{-x,d}^{(\cdot)}}
\end{eqnarray*}

Here, $\mathbf{z}^*$ and $\mathbf{z}'$ denote all the topics governed by the
current sentence and the others, respectively (i.e., $\mathbf{z} = \mathbf{z}^*
\cup \mathbf{z}'$).  We use $n_x^{(z)}$ to denote the count of words in the
current sentences being assigned to topic $z$, and $n_{-x,d}^{(l)}$ to denote
the count of sentences in document $d$ that are of topic $l$, excluding the
current assignment.  Note that $\mathcal{B}(\cdot)$ is the multinomial beta
function defined by \[\mathcal{B}(a_1, \ldots, a_n) = \frac{\prod_i
\Gamma(a_i)}{\Gamma(\sum_i a_i)}. \]

\begin{table*}[!ht]
  \centering
  \begin{tabular}{ll}
    Parameter Equation &  Adjusted $R^2$ \\
    \hline
    $\alpha = 0.6433 \times 1/S$ & 0.4872 \\
    $\beta = 1.46 \times {10^{-4}}^{*} \times S + 1.4528713 \times 1/T$ & 0.877 \\
    $\gamma = 5.276 \times {10^{-5}} \times S + 0.2156 \times 1/T $& 0.9135 
  \end{tabular}

  \caption{The linear fit for model parameters.  All weight values with are
  significant at $p < 0.001$, except $(^*)$
  significant at $p < 0.1$.}\label{t:model-parameters} \end{table*}

\section{Parameter Estimation} \label{s:parameter-estimation} 

The performance of LDA-based models such as this one depends largely on the
parameters for the Dirichlet priors.  To the best of our knowledge, there is no
standard approach for optimizing these parameters; they are normally determined
through experimentation.  Accordingly, we used the following experimental
procedure to estimate the parameters of our model.

We decided to learn the parameters on a large, multi-topic corpus.  We chose to
use the ICWSM 2009 Spinn3r dataset of 44 million weblog posts
\cite{burton2009icwsm} due to its large size and broad coverage over topics.
We took a sample of the corpus and divided it into the training, test, and
development sets.

Our approach was to learn the optimal parameters $(\alpha, \beta, \gamma)$ for
various numbers of word and sentence topics.  To do this, we conducted a
three-dimensional grid search for these parameters when given 5, 10, 50, or 100
word or sentence topics (4 $\times$ 4 possible combinations).  For each
$(\alpha, \beta, \gamma)$ triple, we assigned values of 0.05, 0.01, 0.005, and
0.001 to each parameter and trained the sentence-layered model.  The trained
models were tested against the development set to calculate the test-set
perplexity.  We selected the model with the lowest test-set perplexity and
recorded the corresponding $(\alpha, \beta, \gamma)$ parameters for the given
number of sentence and word topics.

Next, we formed a linear regression model based on the recorded data.  We
derived a linear fit function for each parameter for a given number of
word/sentence topics.  The resulting equations are shown in
Table~\ref{t:model-parameters}.

\begin{figure}[!ht]
  \centering
  \includegraphics[width=0.75\columnwidth]{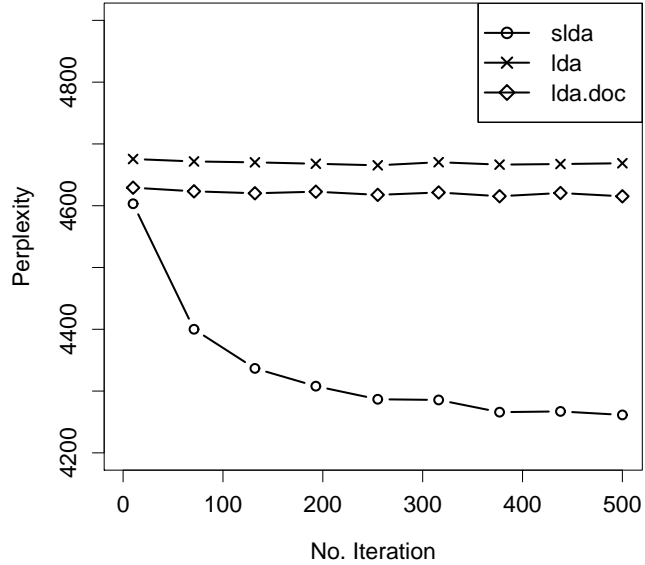}
  \caption{Performance of the three topic models in terms of test-set
  perplexity on the Penn TreeBank.  The sentence-layered LDA is denoted as {\it
  slda}, the original LDA trained on sentence level as {\it lda}, and the
  original LDA trained on document level as {\it lda.doc} }
  \label{f:perplexity}
\end{figure}

\section{Perplexity Evaluation} \label{s:perplexity-evaluation}

One way to assess the performance of a topic model is through test-set
perplexity.  Test-set perplexity evaluates how well the model generalizes on a
held-out data.  Perplexity is measured in the amount of uncertainty, which is a
positive real number.  The lower perplexity we achieve for the model, the
better fit the model is to the data.  The test-set perplexity is calculated as:

\[ \mathrm{perplexity_{test}} = \exp(\frac{- \sum_w \log \Pr(w|d_\mathrm{test})}{N_\mathrm{test}}) \]

The formulation of $\Pr(w_\mathrm{test})$ is different for each topic model.
For standard LDA, the probability is evaluated as follows.  \[
\Pr(w|d_\mathrm{test}) = \sum_z \Pr(w|z) \Pr(z|d) \] The probability for the
sentence LDA is a bit more complicated due to an additional layer of latent
semantics.  \[ \Pr(w|d_\mathrm{test}) = \sum_x \sum_z \Pr(w|z) \Pr(z|x)
\Pr(x|d) \] 

To calculate perplexity for each model, we begin with probabilities $\Pr(w|z)$
learned from the training set.  Since the training and test sets did not share
documents or sentences, the probabilities related to $z$, $x$, and $d$ needed
to be re-sampled using texts from the test data.  We employed the same Gibbs
sampling procedure with the initialized values for $\Pr(w|z)$ and started
another 500-iteration burn-in on the test sets.  For simplicity, we read out
only one sample for $\Pr(z|x)$, $\Pr(x|d)$, and $\Pr(z|d)$ at the end of
burn-in.

As a corpus for evaluation, we chose the documents in the Penn TreeBank
\cite{marcus1994building}.  The Penn TreeBank is composed of 2,312 parsed
documents, where each document contains 21.3 sentences on average.  We further
divided the TreeBank into two sets, one of 2,300 documents and the other of
154.  We trained the topic models on the first set and evaluated the perplexity
on the second.  For both models, we had the Gibbs sampler burned in for 1,000
iterations before reading out probability estimates.  

We set up the parameters of the standard LDA model as suggested in
\cite{griffiths2004finding}, i.e., $\alpha = 50 / T$ and $\beta = 0.01$.  For
the sentence-layered LDA model, the parameters were determined by applying the
regression results described in Section~\ref{s:parameter-estimation}.  In this
experiment, we tested a total of three topic models: our sentence-layered LDA model and
two standard LDA models, one trained at the document level and the other at the
sentence level.  When trained at the document level, LDA discards sentence
boundaries and treats the entire document as one text unit.  When trained at
the sentence level, LDA treat each sentence as individual ``documents''.  Tests
were made only at the sentence level.  

The number of word topics $T$ for all models were set to 10.  The word topics in
both models are functionally equivalent and therefore should be set to the same
value.  The number of sentence topics for the sentence-layered model were
assigned to 20, which empirically achieved the best performance in our
preliminary tests.

The experimental results showed that our sentence-layered LDA model achieved
lower test-set perplexity than the standard LDA models.  As shown in
Figure~\ref{f:perplexity}, the perplexity for our model converged in roughly
300 iterations, while the LDA models were stable at an early stage.  We
believe that both of the standard LDA models suffered from one of two issues.
When the LDA model was applied at the sentence level, the association of
topic words did not span across the sentence boundaries, resulting in a loss of
co-occurrence statistics.  Likewise, when the LDA model was trained at the
document level, it disregarded the structure of the text and treated all the
word occurrences as equally important.  As a result, the model overfitted the
data and failed to address the diversity of co-occurred words at the sentence
level.  Our sentence-layered LDA model avoids these problems by adding a
minimal layer of document structure directly into the model, allowing for a
balance of inferred latent semantics between the document level and the
sentence level.

\section{Related Work} \label{s:related-work} 

This work was inspired by a number of efforts for extending the standard LDA
model.  Our work differs from Pachinko allocation model \cite{li2006pachinko},
nested Chinese restaurant process \cite{blei2004hierarchical}, and mixture
network \cite{heinrich2009generic}, each of which allows an arbitrary number of
sub-document units and an arbitrary number of dependency links in between to
model topic correlations.  Instead, our model adheres strictly to the sentence
boundaries to define document structure.  Our approach is most similar to that
of the latent-Dirichlet co-clustering model \cite{shafiei2006latent}.  Our work
differs in that it utilizes multiple sentence-level LDA machines to better
account for the contribution of sentences to the topics of documents.

\section{Conclusions} \label{s:discussion} 

Although these perplexity results are encouraging, there remains difficult
challenges in directly applying the topic distributions of our model in
sentence-level tasks.  Because there are two topic distributions in our model,
some additional consideration is needed to coordinate the contribution of each
to the generation of words.  Additionally, good perplexity results on their own
do not guarantee that the model captures the features most needed in topic
  modeling tasks.  These challenges will be the focus of our future work.


\bibliographystyle{plain}
\bibliography{report}
\end{document}